%% file: sample-sigconf-authordraft.tex
\documentclass[sigconf]{acmart}

\renewcommand\footnotetextcopyrightpermission[1]{} 
\AtBeginDocument{%
  }

\acmISBN{978-1-4503-XXXX-X/2018/06}

\usepackage{hyperref}
\usepackage{url}
\usepackage{algorithm}
\usepackage{algorithmic}
\usepackage{booktabs}
\usepackage{amsmath}
\usepackage{graphicx}
\usepackage{multirow}
\usepackage{float} 
\usepackage{xcolor}
\usepackage{tcolorbox}
\usepackage{makecell}

\begin{document}

\title{Instruction-aware User Embedding via Synergistic Language and Representation Modeling}

\author{Ziyi Gao}
\authornote{Both authors contributed equally to this research.}
\affiliation{%
  \institution{Ant Group}
  \city{Hangzhou}
  \country{China}
}
\email{yinan.gzy@antgroup.com}

\author{Yike Xu$^{\ast}$}
\affiliation{%
  \institution{Ant Group}
  \city{Hangzhou}
  \country{China}
}
\email{xuyike.xyk@antgroup.com}

\author{Jiahao Yuan}
\affiliation{%
  \institution{Ant Group}
  \city{Hangzhou}
  \country{China}
}
\email{xiaodeng.yjh@antgroup.com}

\author{Baokun Wang}
\authornote{Corresponding author.}
\affiliation{%
  \institution{Ant Group}
  \city{Hangzhou}
  \country{China}
}
\email{yike.wbk@antgroup.com}

\author{Jinyong Wen}
\affiliation{%
  \institution{Ant Group}
  \city{Hangzhou}
  \country{China}
}
\email{wenjinyong.wjy@antgroup.com}

\author{Xiaotong Lin}
\affiliation{%
  \institution{Ant Group}
  \city{Hangzhou}
  \country{China}
}
\email{lxt203095@antgroup.com}

\author{Yun Liu}
\affiliation{%
  \institution{Ant Group}
  \city{Hangzhou}
  \country{China}
}
\email{ly319278@antgroup.com}

\author{Xing Fu}
\affiliation{%
  \institution{Ant Group}
  \city{Hangzhou}
  \country{China}
}
\email{zicai.fx@antgroup.com}

\author{Yu Cheng}
\affiliation{%
  \institution{Ant Group}
  \city{Hangzhou}
  \country{China}
}
\email{cy122623@antgroup.com}

\author{Yongchao Liu}
\affiliation{%
  \institution{Ant Group}
  \city{Hangzhou}
  \country{China}
}
\email{yongchao.ly@antgroup.com}

\author{Weiqiang Wang}
\affiliation{%
  \institution{Ant Group}
  \city{Hangzhou}
  \country{China}
}
\email{weiqiang.wwq@antgroup.com}

\author{Zhongle Xie}
\affiliation{%
  \institution{Zhejiang University}
  \city{Hangzhou}
  \country{China}
}
\email{xiezl@zjut.edu.cn}

\renewcommand{\shortauthors}{Ziyi Gao et al.}

\begin{abstract}
  User representation modeling has become increasingly crucial for personalized applications, yet existing approaches struggle with generalizability across domains and sensitivity to noisy behavioral signals. We present InstructUE, an instruction-aware user embedding foundation model that leverages large language models (LLMs) to generate general and instruction-aware user representations. InstructUE introduces a multi-encoder architecture with a lightweight adapter that efficiently processes heterogeneous data from six different sources while preserving their structural characteristics. Additionally, it proposes a novel contrastive-autoregressive training framework that bridges language and representation spaces through a curated UserQA dataset. The contrastive-autoregressive training framework simultaneously leverages autoregressive learning to capture domain knowledge in language space and contrastive learning to align user-text embeddings in representation space, thereby enhancing the instruction-awareness and noise-robustness of user embeddings. Through extensive experiments on real-world applications, we demonstrate that InstructUE significantly outperforms existing methods across multiple domains including user prediction, marketing, and recommendation scenarios. Our results show that instruction-aware user modeling can effectively achieve instruction-guided denoising of user information in specific scenarios, paving the way for more generalizable and robust user representation learning.
\end{abstract}



\received{20 February 2007}
\received[revised]{12 March 2009}
\received[accepted]{5 June 2009}

\maketitle
\input{sections/1-intro}
\input{sections/2-related}

\input{sections/3-method}
\input{sections/4-exp}

\input{sections/5-conlusion}
\bibliographystyle{ACM-Reference-Format}
\bibliography{sample-base}

\appendix
\section{Implementation Details. }
We fine-tune the InstructUE using a global batch size of 1024 for 150k steps. During contrastive-autoregressive training, LoRA~\cite{hu2022lora} is used in the backbone and the optimizer is AdamW, with cosine decay learning rate initialized at 2e-4, and the embedding sizes of user representation are set to 896. We set $\alpha_{lm}$ and $\alpha_{cl}$ to 1. The model is pre-trained on 128 PPU GPUs, and a single GPU is used for testing.

\section{Baselines. }
We compare InstructUE with a comprehensive set of baseline models, categorized as follows:
\begin{itemize}
    \item \textbf{General LLM Embeddings:} We use general-purpose text embedding models to encode user data, which is formatted as natural language text. \textbf{Qwen2.5-0.5B-Instruct} is the base LLM used without user modeling fine-tuning, for which the last token's output is used as its representation. \textbf{Qwen3-embedding-0.6B}~\cite{zhang2025qwen3} is a state-of-the-art text embedding model.
    \item \textbf{Traditional User Modeling:} \textbf{U-MLP One4all}~\cite{shin2021one4all} is a user targeting model that extends the general-purpose One4all representation by adding an MLP decoder. \textbf{MSDP}~\cite{fu2023robust} and \textbf{CPC}~\cite{oord2018representation} are contrastive learning-based methods that learn from augmented views of user behavior sequences.
    \item \textbf{LLM-based User Representation Models:} \textbf{FOUND}~\cite{dou2025transferable} and SUPERMOE~\cite{jiang2022learning} is recent foundation model designed for user representation, leveraging large-scale data and pre-trained model architectures.
\end{itemize}









\end{document}

%% file: sections/1-intro.tex
\section{Introduction}

User representation modeling learns expressive user embeddings from large-scale, multi-source data such as textual profiles, interaction histories, and tabular attributes, to support accurate user understanding for data-driven, human-centric applications in digital marketing~\cite{zhang2024scaling}, recommendation~\cite{feng2025long}, and financial services~\cite{dou2025transferable}.
To capture richer patterns in historical behavior, prior studies have adopted contrastive learning by aligning augmented views of user behavior via self-supervised objectives such as contextual consistency within behavioral sequences \cite{lin2022dual} and cross-view consistency \cite{zou2022multi}. 
However, as behaviors grow increasingly sequential and context-sensitive, these methods face limitations in modeling complex behavioral dependencies and semantic relationships. Effective representations must not only capture historical patterns but also generalize beyond observed interactions to anticipate future actions~\cite{dou2025transferable}, requiring more sophisticated sequence modeling capabilities compared to traditional approaches.

To this end, recent research has leveraged Language Models (LMs) for user representation, either by fine-tuning on tokenized behavior sequences~\cite{sun2019bert4rec,ning2025user,shestov2025llm4es} or by enriching item semantics using LMs~\cite{hou2022towards}. While such approaches demonstrate the potential of LMs to capture sequential and semantic patterns, they face a fundamental modality mismatch: \textit{user behaviors are sparse, symbolic sequences that differ significantly from the dense natural language used to pretrain LMs.} This mismatch limits the ability to effectively leverage LMs' rich semantic knowledge and flexible reasoning capabilities.
Consequently, current methods face two key challenges: (i) \textbf{Limited generalizability and insufficient open-world knowledge:} domain-specific models like BERT4Rec~\cite{sun2019bert4rec} and RecBole~\cite{zhao2021recbole} excel in e-commerce but generalize poorly across domains such as security. Without access to broad semantic knowledge, they fail to handle complex scenarios such as financial anomalies or promotional campaigns~\cite{teterwak2025large,wei2024llmrec}. (ii) \textbf{Poor flexibility and noise sensitivity:} existing models typically produce static embeddings that do not adapt to downstream instructions. In the absence of explicit guidance, they are easily influenced by noisy or irrelevant behaviors, undermining robustness in real-world applications~\cite{qiu2022contrastive,he2025llm2rec}. Moreover, the quality of task instructions plays a critical role in guiding user representation learning: imprecise instructions often result in degraded embeddings and suboptimal downstream reasoning performance.

To address above limitations, we propose \textbf{InstructUE}, an \textbf{Instruct}ion-aware \textbf{U}ser \textbf{E}mbedding Foundation model. Built upon Large Language Models(LLMs), InstructUE leverages their rich open-world knowledge and semantic reasoning capabilities to construct generalizable user representations. Our key innovation lies in rethinking user representation as an instruction-conditioned process, where natural language instructions can flexibly guide the embedding construction. Specifically, InstructUE encodes six heterogeneous data sources with modality-specific encoders and projects them into the LLM space via lightweight adapters, thereby preserving structural fidelity and enabling scalable source integration while mitigating modality conflicts and context overflow from naïve concatenation. Together with instruction guidance, this design effectively bridges the gap between symbolic user behaviors and semantic understanding.


Furthermore, we bridge language and representation spaces by curating a large-scale, domain-agnostic UserQA dataset and proposing a contrastive–autoregressive finetuning strategy: the autoregressive objective injects domain knowledge in the language space, while contrastive learning aligns users with text in the representation space to perform a form of instruction-guided denoising~\cite{qiu2022contrastive,he2025llm2rec}. Building on the observation that semantically similar texts yield consistent instruction responses~\cite{wei2022finetuned,longpre2023flan}, we extend this principle to user modeling by assuming that users with similar attributes should elicit similar responses under identical instructions, thereby enhancing embedding quality. In addition, InstructUE employs few-shot cluster-based supervised instruction tuning to efficiently adapt embeddings to diverse downstream personalization tasks, where improving the quality of instructions is crucial for enhancing downstream reasoning and embedding transferability.  Our key contributions are summarized as follows:

\begin{figure}[t]
\centering
\includegraphics[width=\linewidth]{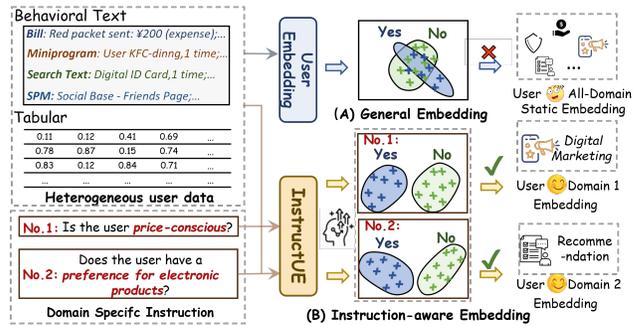} 
\caption{Comparison between (A) General User Embedding \cite{dou2025transferable} and (B) our InstructUE. (A) learns transferable user representations across domains but generates fixed embeddings regardless of downstream context. (B) extends (A) with instruction-aware modulation, enabling a single model to produce adaptive, domain-specific embeddings via natural language instructions.}
\label{Fig.frame}
\end{figure}

\begin{itemize}
    \item We propose InstructUE, an industrial-scale, LLM-based user embedding foundation model capable of understanding users across diverse industry domains via pretraining on large-scale heterogeneous multi-source data.
    \item We introduce a contrastive-autoregressive joint training approach that bridges language and representation spaces to mitigate noise sensitivity, further exploring few-shot instruction tuning that refines several instruction tokens to enhance downstream transferability.
    \item Extensive experiments reveal that InstructUE surpasses baseline methods across diverse scenarios and multiple domains, demonstrating its superior transferability and generalizability in real-world applications.
\end{itemize}

%% file: sections/2-related.tex
\section{Related Work}

\paragraph{Text Embedding.} Text embedding has been a core task in NLP, with early supervised approaches training BERT-like models on sentence-level objectives such as natural language inference and semantic similarity. Contrastive methods such as SimCSE~\cite{gao2021simcse} further showed that simple augmentation can produce high-quality embeddings without labeled data. More recently, decoder-only LLMs have been adapted as embedders. Prior work has commonly used the hidden state of the final token as the sentence representation~\cite{ma2024fine,wang2024improving}, while LLM2Vec~\cite{behnamghader2024llm2vec} augments this with masked prediction and contrastive objectives to enable bidirectional attention. InBedder~\cite{peng2024answer} exploits the generative capacity of LLMs by fine-tuning on abstractive question answering and deriving embeddings from generated answers, and Qwen3-embedding~\cite{zhang2025qwen3} develops a multilingual embedding and reranking model through large-scale unsupervised pretraining and supervised fine-tuning. These works highlight the potential of instruction-driven and generative paradigms to enhance the transferability of embeddings, providing inspiration for user representation learning.

\paragraph{User Representation Learning.} Beyond the text domain, user representation learning seeks to encode heterogeneous behavioral data into compact embeddings. Early frameworks such as DUPN~\cite{ni2018perceive} and MSDP~\cite{fu2023robust} captured sequential dependencies through recurrent and predictive objectives, while contrastive methods improved robustness by aligning augmented user histories across contexts~\cite{lin2022dual,zou2022multi}. More recently, foundation models such as FOUND~\cite{dou2025transferable} attempt to learn general-purpose user embeddings that transfer across domains while remaining predictive of future actions. However, these approaches still assume a static representation per user (or per scenario), limiting flexibility when adapting to new downstream objectives. In contrast, our InstructUE introduces instruction-aware modeling, allowing user embeddings to dynamically adapt to downstream tasks and scenario-specific personalization.

\paragraph{Language Models for User Modeling.}  
Recent studies have increasingly employed large language models (LLMs) for user modeling. Transformer-based approaches such as BERT4Rec~\cite{sun2019bert4rec} and its LLM extensions~\cite{ning2025user,shestov2025llm4es} adapt pretrained models to behavioral sequences to capture temporal dependencies. Alternatively, semantic-enrichment methods exploit textual item descriptions to enhance user embeddings and improve cross-task generalization~\cite{hou2022towards,wei2024llmrec}. Despite these advances, a fundamental modality mismatch remains: user behaviors are sparse and symbolic, in contrast to the dense natural language on which LLMs are pretrained. Consequently, most methods rely on shallow prediction objectives such as next-item or masked-action modeling~\cite{nguyen2025jepa4rec,du2023ensemble}, and domain-specific frameworks (e.g., RecBole~\cite{zhao2021recbole}) exhibit limited transferability beyond narrow application scenarios~\cite{teterwak2025large}. Moreover, static embeddings remain highly sensitive to noisy or irrelevant behaviors~\cite{he2025llm2rec}, undermining robustness in real-world environments. To overcome these challenges, we propose InstructUE, which leverages LLMs in an instruction-conditioned process and introduces an autoregressive–contrastive strategy to inject semantic knowledge and suppress noisy behaviors, yielding robust and adaptable user embeddings.

%% file: sections/3-method.tex
\section{Method}
\label{sec:method}

\begin{figure*}[t]
\centering
\includegraphics[width=\textwidth]{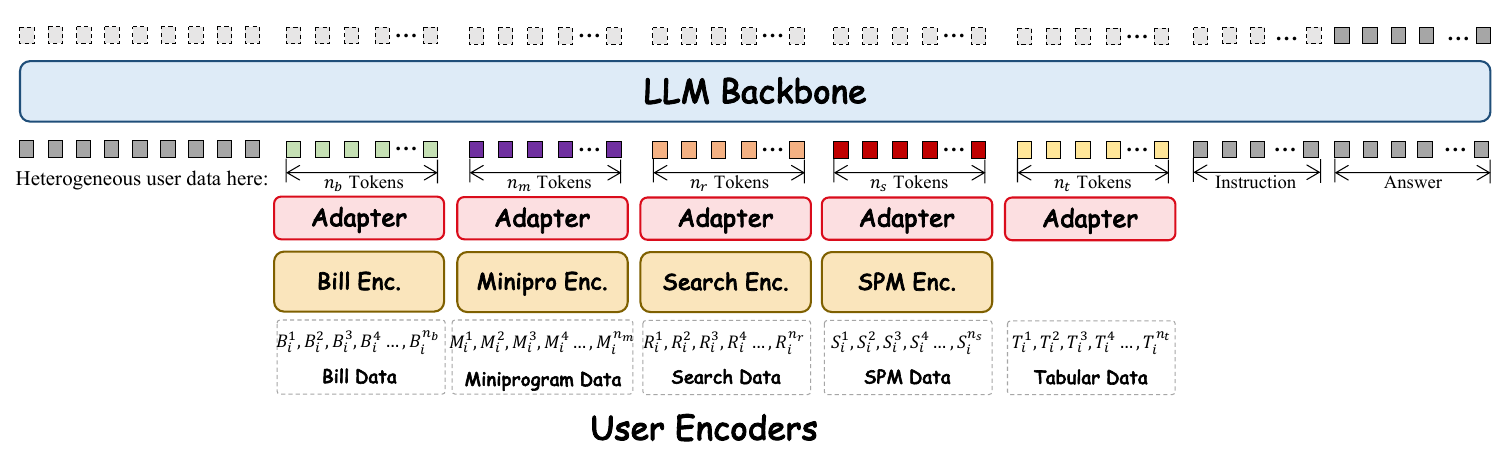} 
\caption{Overview of the framework. Multi-source user data is formatted into natural language sequences with modality delimiters. Instructions guide the generation process, and the \texttt{<USER>} token extracts the final embedding.}
\label{Fig.frame_detail}
\end{figure*}

Existing user representation models typically aim to produce a single, static embedding that captures a holistic view of a user’s behavior \cite{dou2025transferable}. While effective for general-purpose retrieval or clustering, such representations lack flexibility when downstream tasks demand different semantic interpretations of the same user, such as predicting spending habits versus inferring lifestyle preferences. This rigidity limits their applicability in real-world scenarios where the same user should be represented differently depending on the task-specific instruction. 

To address the above limitation, we propose \textbf{InstructUE}, an instruction-aware user representation learning framework that enables instruction-guided embedding generation. Through training on our synthetic UserQA dataset (Sec.~\ref{sec:userqa}) and leveraging synergistic language and representation learning (Sec.~\ref{sec:synergistic}), InstructUE learns to dynamically adapt user representations based on task-specific directives. This allows the same user to be represented differently depending on the downstream objective, all within a unified model architecture.

\subsection{Problem Formulation}
\label{subsec:formulation}
We cast instruction-aware user representation learning as a conditional embedding task, where the goal is to map a user's multi-modal behavioral profile $X_i = \{B_i, M_i, S_i, \mathcal{R}_i, T_i\} \in \mathcal{X}$ and a natural language instruction $I_k \in \mathcal{I}$ into a dense vector $u_{i,k} = f(X_i, I_k) \in \mathbb{R}^d$ that encodes only those aspects of behavior semantically relevant to $I_k$. Here, $i \in \{1, \dots, N\}$ indexes users, $B_i = \{B_i^1, \dots, B_i^{n_b}\}$ denotes sequences of PayBill transactions, $M_i$ captures Mini Program interactions, $S_i$ represents super position model (SPM) paths, $\mathcal{R}_i$ records homepage search queries—all sequential modalities—while $T_i \in \mathbb{R}^{F \times D}$ aggregates $F$ tabular features with $D$-dimensional embeddings. The function $f: \mathcal{X} \times \mathcal{I} \to \mathbb{R}^d$ thus generates dynamic, intent-conditioned representations such that the same user $i$ can be embedded differently under instructions like “Identify this user’s preferred dining categories” or “Estimate next month’s food delivery spend,” enabling fine-grained semantic steering without architectural modification. In the absence of an instruction, $f$ reduces to a holistic encoder $u_i = f(X_i)$, producing a general-purpose representation by fusing cross-modal patterns into a unified view. This dual-mode paradigm—conditional when guided, integrative when free—endows the model with zero-shot adaptability across diverse downstream tasks, treating natural language instructions as soft controllers over the latent space and unifying flexible user modeling with precise, semantics-driven representation learning.

\subsection{UserQA: Synthetic Instruction-Response Data Construction}
\label{sec:userqa}

\paragraph{User Source Data.}
To enable both general and instruction-specific understanding, we construct a synthetic training dataset \textsc{UserQA} via large language model (LLM) driven data synthesis. Each instance is a triplet $(X_i, I_k, A_k)$, where $X_i$ denotes the user's multi-source profile, $I_k \in \mathcal{I}$ is a natural language instruction, and $A_k$ is the expected answer generated from $X_i$ under $I_k$. The input comprises heterogeneous behavioral sequences and structured tabular attributes. Behavioral modalities include PayBill information ($B$), Mini Program logs ($M$), SPM details ($S$), and search text ($\mathcal{R}$). Each modality is transformed into fluent natural language using domain-specific templates; for example, billing records are rendered as descriptive narratives incorporating transaction items, categories, amounts, and payment channels. Tabular data is represented as a tensor $T \in \mathbb{R}^{N \times F \times D}$, where $N$ is the number of users, $F$ the feature count, and $D$ the dimensionality of each embedded feature. Thus, each user's full profile is defined as $X_i = \{B_i, M_i, S_i, \mathcal{R}_i, T_i\} \in \mathcal{X}$, with $i \in \{1, \dots, N\}$ and $\mathcal{X}$ denoting the joint space of all possible user data.

\begin{figure}[t]
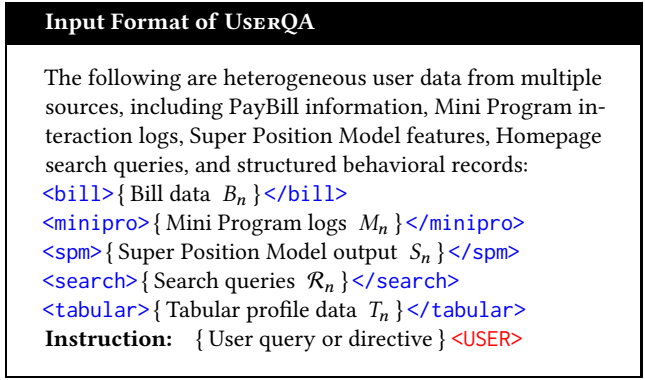

\centering
\begin{tcolorbox}[
    width=\linewidth,
    colback=white,
    colframe=black,
    title=Input Format of \textsc{UserQA},
    fonttitle=\bfseries\normalsize,
    sharp corners,
    boxrule=1pt,
    top=6pt,
    bottom=6pt
]
The following are heterogeneous user data from multiple sources, including PayBill information, Mini Program interaction logs, Super Position Model features, Homepage search queries, and structured behavioral records:

\textcolor{blue}{\texttt{<bill>}}\,\{ \textnormal{Bill data } $B_n$ \}\,\textcolor{blue}{\texttt{</bill>}}

\noindent
\textcolor{blue}{\texttt{<minipro>}}\,\{ \textnormal{Mini Program logs } $M_n$ \}\,\textcolor{blue}{\texttt{</minipro>}}

\noindent
\textcolor{blue}{\texttt{<spm>}}\,\{ \textnormal{Super Position Model output } $S_n$ \}\,\textcolor{blue}{\texttt{</spm>}}

\noindent
\textcolor{blue}{\texttt{<search>}}\,\{ \textnormal{Search queries } $\mathcal{R}_n$ \}\,\textcolor{blue}{\texttt{</search>}}

\noindent
\textcolor{blue}{\texttt{<tabular>}}\,\{ \textnormal{Tabular profile data } $T_n$ \}\,\textcolor{blue}{\texttt{</tabular>}}

\noindent
\textbf{Instruction:}\quad \{ \textnormal{User query or directive} \} \textcolor{red}{\texttt{<USER>}}
\end{tcolorbox}
\caption{
Input format of \textsc{UserQA}. Each modality is wrapped in semantic delimiters (shown in \textcolor{blue}{blue}). 
An optional instruction is followed by the special \textcolor{red}{\texttt{<USER>}} token, which signals the model to extract a unified user embedding.
}
\label{fig:input_format}
\end{figure}

As illustrated in Figure~\ref{fig:input_format}, model inputs support two modes: instruction-aware and general. In the instruction-aware mode, the instruction $I_k$ is appended to the user data sequence; for general embeddings, no instruction is provided, enabling task-agnostic representation learning. We adopt a causal LLM and append a special \texttt{<USER>} token at the end of the input sequence. The final-layer hidden state corresponding to this token is used as the user embedding. To preserve modality structure, each data source is enclosed with boundary tokens such as \texttt{<bill>} and \texttt{</bill>}. During autoregressive training, we employ teacher forcing by appending the ground truth answer to the full input, allowing the model to learn accurate response generation conditioned on both the user profile and the instruction.

\paragraph{Synthesized Training Data.}
We construct \textsc{UserQA}, a dataset designed to train both general and instruction-aware user representation models, by synthesizing data from a real-world industrial environment along two complementary axes. First, we mine stable user attributes from historical multi-source data. An LLM analyzes behavioral traces and generates instruction-answer pairs that reflect inferred characteristics. For instance, given frequent high-end dining transactions in PayBill records, the LLM may produce: ``What is this user's dining preference?'' $\to$ ``The user prefers high-end dining establishments.'' This encourages the model to form semantically meaningful, static user profiles from complex inputs.

Second, we construct forward-looking, business-relevant instructions grounded in users' future behaviors. For example, based on observed food delivery patterns, we prompt: ``Please predict the user's detailed food delivery expenses for the next month.'' The LLM generates plausible forecasts informed by temporal trends and contextual signals. This trains the model to capture dynamic behavioral patterns and produce predictive embeddings useful for downstream applications such as spend forecasting or personalized recommendation.

All synthesized instances follow the triplet structure $(X_i, I_k, A_k)$, where $X_i$ contains the user's multi-source data up to the current time, $I_k$ specifies the task, and $A_k$ provides the target response. For general representation learning, we set $I_k = \emptyset$, prompting the model to encode the full profile without task guidance, effectively encouraging comprehensive understanding.

This dual-path synthesis strategy ensures that the model learns both broad user understanding for open-ended tasks and precise, instruction-following capabilities for targeted applications. By covering descriptive, diagnostic, and predictive scenarios, \textsc{UserQA} fosters robust, versatile, and application-ready user representations.

\subsection{InstructUE: Synergistic Language and Representation Learning}
\label{sec:synergistic}

\subsubsection{Model Architecture}

Following the design of MLLM~\cite{wang2024qwen2,liu2023visual}, the overall network architecture of InstructUE is illustrated in Fig.~\ref{Fig.frame_detail}. In terms of language processing, InstructUE adopts the pre-trained LLM backbone as its foundation component. Considering the long-range contextual dependencies in users' sequential textual behaviors, we encode each textual data source as user features for input processing. Specifically, each textual user data source is independently processed through a user encoder and an adapter, while numerical tabular data is processed through the adapter. The user encoder of InstructUE uses the BERT-based model, and a single-layer MLP serves as the adapter.

\begin{figure*}[t]  
\centering
\includegraphics[width=\textwidth]{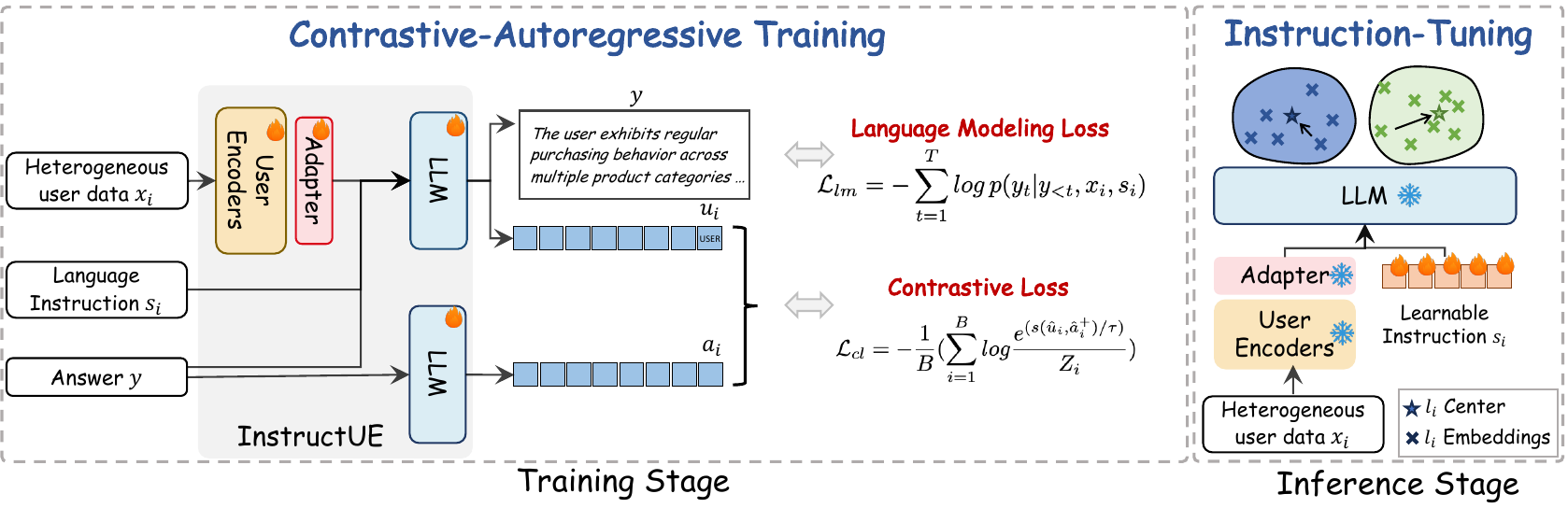} 
\caption{Training and inference pipeline of InstructUE. User representations are learned via a contrastive-autoregressive joint strategy, while instruction quality is enhanced through cluster-based supervised tuning of learnable instructions.} 
\label{Fig.frame} 
\end{figure*}

\subsubsection{Comtrastive-Autoregressive Strategy}
Although LLMs demonstrate superior generative abilities through autoregressive training paradigms, they suffer from inherent limitations in discriminative representation capabilities. To effectively learn user representations that capture user profiling features from user data, we propose a Comtrastive-Autoregressive training framework that combines contrastive learning with autoregressive training. 

\textbf{Autoregressive Language Learning.}
Following the standard approach in LLM training, we train the model in a teacher-forcing strategy to predict the next token based on the previous tokens. The autoregressive language learning objective is as follows:

\begin{equation}
\mathcal{L}_{lm} = -\sum_{t=1}^T log\,p(y_t|y_{<t},x_i,s_i).
\end{equation}
where $y_t$ is the target $t$-th token and $p(y_t|y_{<t},x_i,s_i)$ is the predicted probability distribution over the vocabulary.

As demonstrated in ~\ref{sec:exp_abl}, the autoregressive training approach inherently enhances model representation capabilities. This enhancement can be understood through two key mechanisms:
Firstly, the autoregressive training preserves the LLM's inherent generative capabilities, enabling it to acquire domain-specific user understanding through the prediction of token distributions in responses.
Secondly, previous research~\cite{peng2024answer} has demonstrated that texts sharing similar implicit semantics tend to generate comparable responses following the same instruction in LLM-based text representations, thus leading to more similar embeddings. Extending to user modeling, similar user behaviors naturally elicit analogous responses under identical instructions. This autoregressive objective effectively captures embedded semantic features through response generation, simultaneously improving the interpretability of the embeddings. 

\textbf{Contrastive Learning.} 
To learn user representations, we align the user embeddings $u_i$ with answer embeddings $a_i$ to user embedding. The training objective is implemented via InfoNCE loss, which guides the model to learn discriminative representations by attracting semantically related samples and repelling negative samples in the embedding space. 
The alignment mechanism between user representations and answer-encoded attributes and behaviors facilitates the clustering of users exhibiting similar behavioral patterns in the instruction-guided representation space. This approach serves two fundamental objectives: (1) extracting comprehensive user profiles from historical behavioral sequences, and (2) developing predictive capabilities through the learning of future behavior patterns.

The contrastive learning objective can be formulated as follows:
\begin{equation}
\mathcal{L}_{cl}=-\frac{1}{B}(\sum_{i=1}^B log\frac{e^{s(\hat{u}_i,\hat{a}_i^+)/\tau}}{Z_i}),
\end{equation}
where $\hat{u}_i,\hat{a}_i$ is normalized user and answer embeddings, $s(\cdot,\cdot)$ denotes the similarity function (we adopt cosine similarity), and $\tau \in \mathbb{R}^+$ is the temperature parameter, and $Z_i$ is the normalization factor that aggregates the similarity scores of the positive pair against negative pairs. Inspired by ~\cite{zhang2025qwen3}, we add same-side representations to the negative pairs:
\begin{align}
Z_i ={} & e^{s(\hat{u}_i,\hat{a}_i^{+})/\tau} 
+ \sum_{j \neq i} m_{ij} e^{s(\hat{u}_i,\hat{a}_j)/\tau} \nonumber \\
& + \sum_{j \neq i} m_{ij} e^{s(\hat{u}_i,\hat{u}_j)/\tau} 
+ \sum_{j \neq i} m_{ij} e^{s(\hat{a}_i,\hat{a}_j)/\tau},
\end{align}

where $a_i^+$,$a_j$ is the positive and other in-batch answer embeddings, $u_j$ is other in-batch user embeddings. And we use mask factor $m_{ij}$ to mitigate the impact of false negative samples:
\begin{equation}
m_{ij} = 
\begin{cases}
    0 & \text{if } s_{ij} > s(\hat{u}_i, \hat{a}_i^+) + c_{margin}, \\
    1 & \text{otherwise}.
\end{cases}
\end{equation}

among which $s_{ij}$ is the corresponding score of $\hat{u}_i,\hat{u}_j$ or $\hat{u}_i,\hat{a}_j$ and $c_{margin}$ is a pre-defined margin hyperparameter.

These same-side negative samples enhance the distinctiveness between user representations under different instructions and the distinctiveness of answer representation spaces.

\textbf{Contrastive-Autoregressive Joint Training.} The overall objective function combines the autoregressive language learning and contrastive learning objectives:
\begin{equation}
\mathcal{L}=\alpha_{lm}\mathcal{L}_{lm}+\alpha_{cl}\mathcal{L}_{cl}.
\end{equation}

The autoregressive learning and contrastive learning demonstrate mutual enhancement in representation modeling. Autoregressive learning improves sequence-level semantic understanding and generation capabilities by predicting the next token, enabling the prediction of answer distributions through user embeddings. Contrastive learning, through the comparison of positive and negative samples, focuses on instance-level user semantic similarities, causing similar representations to cluster while dissimilar representations disperse under the same instruction, thereby enhancing the discriminative power of representations. These two optimization objectives complement each other at different granularities, jointly constructing more generalizable user understanding.

\subsubsection{Instruction Precision Matters}
The effectiveness of LLM-based user representations heavily depends on the quality of instructions used to guide their generation. Our experiments in Section~\ref{sec:exp_instruction} show that imprecise instructions often lead to suboptimal representations, providing limited benefits for downstream tasks.
The semantic gap between model's understanding and actual business scenarios can lead to suboptimal clustering in the representation space. This is particularly evident in cases such as product preference definition, where model interpretations may not align perfectly with real-world business contexts.

To mitigate this issue, we introduce a few-shot cluster-based instruction-tuning framework that leverages supervised contrastive learning with downstream labels $l_i$. Specifically, we optimize the instruction tokens by maximizing the similarity between user embeddings within the same class while minimizing it across different classes. This is formulated as a contrastive loss $\mathcal{L}_{pt}$ where user embeddings $\hat{u}_i$ are pulled closer to their corresponding class center $\hat{p}_{l_i}$ (computed as the mean of all user embeddings in that class) while being pushed away from prototypes of other classes, effectively embedding domain-specific business knowledge into instruction tokens. The optimized function can be formulated as follows:

\begin{equation}
\mathcal{L}_{\text{pt}} = -\frac{1}{B} \sum_{i=1}^B 
\log \left( \frac{e^{s(\hat{u}_i, \hat{p}_{l_i}) / \tau}}
{\sum_{k=1}^K e^{s(\hat{u}_i, \hat{p}_k) / \tau}} \right),
\quad \text{where} \quad
p_k = \frac{1}{|C_k|} \sum_{i \in C_k} u_i.
\end{equation}
where $K$ represents the total number of classes in the downstream task, $\hat{p}_k$ is the normalized prototype embedding for class $k$, $\hat{p}_{l_i}$ is the prototype for the correct class of user $i$ and $C_k$ is the set of all users belonging to class $k$.

%% file: sections/4-exp.tex
\section{Experiments}
In this section, we conduct extensive experiments on our proposed InstructUE to demonstrate its superior generalization across diverse domains, validate the effectiveness of its instruction-aware guidance, and dissect the contribution of each core architectural and training component.

\subsection{Experimental Setups}
\noindent\textbf{Datasets.}
InstructUE has been trained on the UserQA dataset compressing a total of 200 million samples.
The user representation test benchmark contains 6 scenarios, covering basic attributes, user prediction, recommendation, and marketing domain, with each scenario having 1 million samples. The test data is in the format of \{$X_i$,$l_i$\} where $X_i$ is heterogeneous multi-source user data and $l_i \in \{0,1\}$ represents whether the user conducts the operations or has the attributes in the specific domain, which is annotated according to real-world business scenario. We employ a linear probing approach to evaluate user representation capability, ensuring that the fitting and evaluation datasets are non-overlapping. For linear probe analysis, each test dataset is split into a 50\% fitting set and an 50\% evaluation set, of which the fitting set is used for linear layer fitting while the evaluation set is used for evaluation. Table ~\ref{tab:my_label} shows the information of our training and test dataset.
\begin{table}[H]
\small
    \centering
    \begin{tabular*}{\linewidth}{@{\extracolsep{\fill}} cccc }
    \hline
         Dataset & No. & Domain & Scenario \\
         \hline
         $\mathcal{D}_{train}$& - & General & General \\
         \hline
         \multirow{6}{*}{$\mathcal{D}_{test}$} & 1 & User Prediction & Ant Forest Click  \\
         & 2 & User Prediction & Game Top-up \\
         & 3 & Basic Attributes & Public Transit  \\
         & 4 & Basic Attributes &  Purchasing Power  \\
         & 5 & \makecell{Recommendation\\\&Marketing} & Preference For Food  \\
         & 6 & \makecell{Recommendation\\\&Marketing} & Preference For Film  \\
         \hline
    \end{tabular*}
    \caption{Data information for user pretraining as well as the test benchmarks.}
    \label{tab:my_label}
\end{table}

\noindent\textbf{Models.}
We adopt Qwen2.5-0.5B-Instruct~\cite{team2024qwen2} as the backbone LLM and gte-base~\cite{li2023towards} as the user multi-source data encoders for our training and inference framework.

\noindent\textbf{Evaluation Metrics. }Linear probe representation analysis is conducted on all annotated datasets, which is evaluated by AUC (Area Under the ROC Curve~\cite{bradley1997use}) and KS (Kolmogorov-Smirnov~\cite{berger2014kolmogorov}) metrics.


\begin{table*}[htbp]
\centering
\caption{Main results on six downstream tasks. Performance is measured by AUC(\%) / KS. The best results are highlighted in \textbf{bold};the second-best are \underline{underlined}. Our proposed methods are listed in the last three rows.}
\resizebox{\textwidth}{!}{
\label{tab:main_results}
\begin{tabular}{lcccccc}
\toprule
\textbf{Models}                & \textbf{Ant Forest Click} & \textbf{Game Top-up} & \textbf{Public Transit} & \textbf{Purchasing Power} & \textbf{Food Preferences }& \textbf{Film Preferences} \\ \midrule
Qwen2.5-0.5B           & 83.49/0.5145 & 94.3/0.7720 & 61.85/0.1657 & 79.84/0.4373 & 74.56/0.3859 & 69.85/0.2940 \\
Qwen3-embedding-0.6B    & 86.08/0.5639 & 86.08/0.7644 & 67.85/0.2623 & 88.12/0.5916 & 76.67/0.4050 & 75.17/0.3698 \\
U-MLP One4all~\cite{shin2021one4all}         & 96.14/0.8191 & 97.31/0.8538 & 62.76/0.1799 & 83.93/0.4926 & 79.84/0.4639 & 75.26/0.3931 \\
MSDP~\cite{fu2023robust}                  & 95.80/0.8042 & 97.11/0.8432 & 63.80/0.1969 & 84.77/0.5016 & 79.09/0.4400 & 76.28/0.3935 \\
CPC~\cite{oord2018representation}                   & 96.08/0.8197 & 97.15/0.8482 & 63.76/0.1946 & 84.15/0.4880 & 80.09/0.4761 & 75.26/0.3839 \\
FOUND~\cite{dou2025transferable}                 & \textbf{97.27}/0.8511 & \underline{98.38}/\underline{0.8951} & 69.44/0.2768 & 94.55/0.7457 & 84.69/0.5465 & 85.44/0.5318 \\ 
SUPERMOE~\cite{jiang2022learning}               & 81.16/0.4732 & 83.23/0.5180 & 63.32/0.1915 & 92.95/0.7138 & 72.92/0.3419 & 77.14/0.4065 \\ \hline
$InstructUE_{Universal}$      & 97.14/0.8467 &97.92/0.8801  & 70.38/0.2910 & \underline{95.57}/\underline{0.7666} & \underline{85.25}/\underline{0.5600} & 85.92/0.5424 \\ 
$InstructUE_{Instruction-manual}$      & 97.10/\textbf{0.8533} & 98.12/0.8887 &\textbf{70.57}/\textbf{0.2935} & 94.50/0.7361 & 83.80/0.5371 & \underline{85.97}/\underline{0.5479} \\ 
$InstructUE_{Instruction-tuned}$      & \underline{97.26}/\underline{0.8524} & \textbf{98.46}/\textbf{0.8972} & \underline{70.50}/\underline{0.2926} & \textbf{95.69}/\textbf{0.7700} & \textbf{85.57}/\textbf{0.5657} & \textbf{86.45}/\textbf{0.5530} \\ \bottomrule
\end{tabular}
}
\end{table*}

\subsection{Main Results and Analysis}
Table~\ref{tab:main_results} presents the main experimental results across all six downstream tasks. The final three rows of Table~\ref{tab:main_results} showcase the performance of our proposed methods, which correspond to three key variants of the InstructUE model deployed at inference:
\begin{itemize}
    \item $InstructUE_{Universal}$: Generates a general-purpose user embedding without any specific instruction.
    \item $InstructUE_{Instruction-manual}$: Generates an embedding conditioned on a hand-crafted, task-relevant instruction.
    \item $InstructUE_{Instruction-tuned}$: Further refines the instruction tokens using few-shot supervised tuning on the downstream task.
\end{itemize}

Notably, Our proposed model, InstructUE, consistently achieves superior or highly competitive performance, demonstrating its strong generalizability and effectiveness. Specifically,our best performance $InstructUE_{Instruction-tuned}$ achieves state-of-the-art results on five out of six tasks. 
Notably, in the film preference scenario, InstructUE surpasses the strong LLM-based baseline, Found, by 1.01\% in AUC. In the `Purchasing Power' and `Food Preferences' domains, it significantly outperforms all baselines, achieving AUC scores of 95.69\% and 85.57\%, respectively. This highlights the model's capacity to integrate rich semantic knowledge from the LLM with fine-grained behavioral patterns from multi-source data.

Furthermore, InstructUE consistently outperforms specialized user representation models like U-MLP and CPC, proving the benefit of leveraging the vast world knowledge and reasoning capabilities inherent in LLMs.

\subsection{The Impact of Instructions}
\label{sec:exp_instruction}
\begin{figure*}[t]  
\centering
\includegraphics[width=0.7\textwidth]{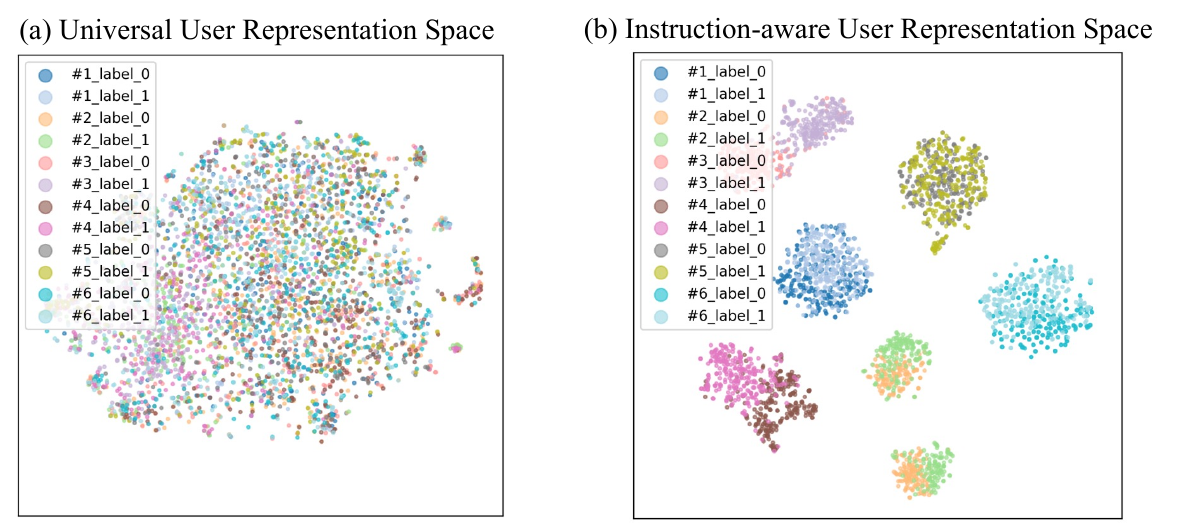} 
\caption{Visualization of universal and instruction-aware representation of six senarios.} 
\label{Fig:cluster} 
\end{figure*}

A comparison between the instruction-guided variants and the $InstructUE_{Universal}$ model reveals the clear benefit of conditioning user representations on instructions. This transformative effect is vividly illustrated in the t-SNE visualization (Figure~\ref{Fig:cluster}). While the universal embeddings form a single, undifferentiated cloud with no discernible structure, the instruction-aware embeddings resolve into six distinct, well-separated clusters, each corresponding to a downstream task. This demonstrates that instructions effectively reorganize the representation space, enhancing intra-class cohesion and inter-class separation to create more discriminative embeddings.

The practical benefit of this improved structure is directly reflected in downstream performance. In tasks like ``Game Top-up,'' ``Public Transit,'' and ``Film Preferences,'' the instruction-aware models consistently outperform their universal counterpart. For instance, in the ``Game Top-up'' domain alone, incorporating a hand-crafted instruction boosts the AUC from 97.92\% to 98.12\%, confirming that these more discriminative, well-clustered embeddings lead to superior task performance.

Interestingly, this positive effect is not guaranteed, especially for tasks with complex or ambiguous business definitions, such as ``Ant Forest Click'' and ``Purchasing Power''. In these scenarios, the manually specified instruction leads to a slight degradation in performance compared to the universal embedding. This suggests that imprecise instructions can sub-optimally narrow the model's focus, inadvertently discarding useful signals and resulting in suboptimal representations. This observation reinforces a key premise of our work: the \textit{quality} and \textit{precision} of the instruction are paramount.

The performance leap from $InstructUE_{Instruction-manual}$ to the $InstructUE_{Instruction-tuned}$ provides compelling evidence for the importance of instruction quality. By fine-tuning the instruction tokens with downstream labels, we bridge the semantic gap between the model's general understanding and the specific business objective. This refinement leads to significant performance gains across most tasks. For example, on the ``Food Preference'' task, AUC improves from 83.80\% to 85.57\%, and on ``Purchasing Power'', it rises from 94.50\% to 95.69\%. These results strongly suggest that finding a better instruction is a highly effective way to enhance downstream performance. The instruction-tuning process enables the model to generate embeddings that are not just task-aware, but are precisely aligned with the nuances of the target application.

\subsection{Ablation Studies}
\label{sec:exp_abl}
We conduct ablation studies to dissect the contributions of the InstructUE's components—input modalities, training methods, and representation selection strategies—with results presented in Table~\ref{tab:ablation_study}.

\begin{table*}[htbp] 
\centering
\caption{Ablation study of InstructUE. The performance is reported in AUC(\%)/KS format. `w/o' means ``without''. The proposed full model, \textbf{InstructUE}, is highlighted in bold.}
\label{tab:ablation_study}
\begin{tabular}{lcccccc}
\toprule
\textbf{Model / Variant} & \textbf{Ant Forest Click} & \textbf{Game Top-up} & \textbf{Public Transit} & \textbf{Purchasing Power} & \textbf{Food Pref.} & \textbf{Film Pref.} \\ 
\midrule
\multicolumn{7}{c}{\textit{Ablation on Modality}} \\
\cmidrule(lr){2-7}
Text Only         & 96.88/0.8300 & 98.16/0.8866 & 69.80/0.2818 & 95.50/0.7643 & 84.35/0.5426 & 85.43/0.5324 \\
w/o Sequence      & 94.95/0.7609 & 96.58/0.8235 & 66.39/0.2273 & 94.64/0.7393 & 81.15/0.4798 & 76.24/0.3913 \\
w/o Tabular       & \textbf{97.16/0.8498} & \textbf{98.18/0.8889} & 69.97/0.2848 & 95.32/0.7539 & 85.12/0.5600 & 85.53/0.5375 \\
\midrule

\multicolumn{7}{c}{\textit{Ablation on Training Method}} \\
\cmidrule(lr){2-7}
w/o NTP             & 96.67/0.8302 & 97.69/0.8697 & 68.38/0.2604 & 95.31/0.7562 & 84.06/0.5379 & 85.01/0.5274 \\
w/o Contrastive   & 96.28/0.8110 & 98.08/0.8844 & 65.64/0.2209 & 93.63/0.7146 & 82.13/0.5024 & 82.80/0.4852 \\
\midrule

\multicolumn{7}{c}{\textit{Ablation on Representation Selection}} \\
\cmidrule(lr){2-7}
Avg. Pooling      & 97.11/0.8456 & 98.11/0.8853 & 69.84/0.2841 & 95.41/0.7618 & 84.92/0.5524 & 84.41/0.5148 \\
\midrule

\textbf{InstructUE (Ours)} & 97.14/0.8467 & 97.92/0.8801 & \textbf{70.38/0.2910} & \textbf{95.57/0.7666} & \textbf{85.25/0.5600} & \textbf{85.92/0.5424} \\ 
\bottomrule
\end{tabular}
\end{table*}

\subsubsection{Ablation on Modality.}
We first investigate the contribution of different input modalities. 
Our dedicated user encoding method proves more effective for sequence data than treating it as unstructured plain text. This is demonstrated by comparing the w/o tabular result, which uses modality-specific encoders, against the text only baseline.  The former consistently outperforms the latter, showing advantages on four out of six tasks, such as ``Food Pref.'' (85.12 vs. 84.35 AUC) and ``Film Pref.'' (85.53 vs. 85.43 AUC).
This demonstrates that our modality-specific encoding is more beneficial for representation learning than simply treating all user data as unstructured plain text. Furthermore, the performance drops significantly when either modality is removed. The w/o Sequence variant, which relies only on tabular data, sees a substantial decrease in performance across all tasks (e.g., a drop from 97.14 to 94.95 AUC on 'Ant Forest Click'). Similarly, removing tabular data (w/o Tabular) also leads to a performance degradation on most tasks compared to the full model. This confirms that both sequential and tabular data provide complementary information, and each modality contributes positively to the model's generalization capability.

\subsubsection{Ablation on Training Method.}
Next, we analyze the impact of our synergistic training framework. The results clearly show that removing either the contrastive loss (w/o Contrastive) or the autoregressive next-token prediction (w/o NTP) leads to a consistent performance drop across all scenarios, demonstrating that contrastive and autoregressive learning are mutually reinforcing. Notably, the model trained only with contrastive learning (w/o NTP) generally outperforms the one trained only with autoregressive learning (w/o Contrastive). For instance, on the ``Purchasing Power' task, w/o NTP achieves a 95.31\% AUC, significantly higher than the 93.63\% from w/o Contrastive. This suggests that while autoregressive training helps inject domain knowledge and improves semantic understanding, the contrastive objective is more direct and effective at learning discriminative representations by explicitly structuring the embedding space. 

\subsubsection{Ablation on Representation Selection.}
Finally, we evaluate the method for extracting the final user representation. The Avg. Pooling variant, which averages the last-layer hidden states of all user and instruction tokens, shows a slight but consistent performance degradation compared to our default approach of using the hidden state of the final <USER> token. This result is expected for decoder-based architectures. The causal attention mechanism is trained to accumulate and synthesize information into the final token's representation for subsequent prediction. The special <USER> token acts as a designated "summary" vector, which the model learns to populate with a holistic, instruction-conditioned representation. In contrast, average pooling treats all input tokens equally, potentially diluting the final representation with less relevant information and failing to capture the synthesized summary that the decoder is designed to produce at its final step. This confirms that for our decoder architecture, using the last token's state is a more effective strategy for extracting a high-quality user embedding.

%% file: sections/5-conlusion.tex
\section{Conclusion}
In this paper, we introduced \textbf{InstructUE}, an instruction-aware user embedding foundation model designed to overcome the rigidity and limited generalizability of traditional methods. InstructUE leverages the semantic power of LLMs to generate dynamic, task-specific user embeddings conditioned on natural language instructions. Our key innovation lies in a synergistic contrastive-autoregressive training strategy, which effectively bridges language and representation modeling to produce robust and discriminative embeddings from heterogeneous data. Extensive experiments validate that InstructUE surpasses state-of-the-art baselines across diverse domains. Our findings further reveal that the quality of instructions is paramount, and that optimizing them via few-shot tuning is a crucial step toward achieving maximal performance on downstream tasks.